\documentclass[10pt]{cai26}
\usepackage{booktabs}
\usepackage{tabularx}
\usepackage{array}
\usepackage{placeins}
\usepackage{hyperref}
\usepackage{algorithmicx}
\usepackage{algorithm}
\usepackage{algpseudocode}

\newcolumntype{Y}{>{\RaggedRight\arraybackslash}X}

\begin{document}
% Editorial staff will replace the following values:
% 1. Conference Year
% 2. Issue number
% 3. Article DOI
\def\conferenceyear{2026}
\volumeheader{39}{0}%{00.000}
\begin{center}

\title{Reason and Verify: A Framework for Faithful Retrieval-Augmented Generation}
\maketitle

\thispagestyle{empty}
\pagenumbering{gobble}

% Add Authors and Affiliations in the camera ready
% for the double blind review, please leave this section as is 
\begin{tabular}{cc}
Eeham Khan\upstairs{\affilone}, Luis Rodriguez\upstairs{\affiltwo}, Marc Queudot\upstairs{\affiltwo,*}
\\[0.25ex]
{\small \upstairs{\affilone} Concordia University, Montréal, QC, Canada} \\
{\small \upstairs{\affiltwo} Centre de recherche informatique de Montréal (CRIM), Montréal, QC, Canada} \\
\end{tabular}
  
% Replace with corresponding author email address
\emails{
  \upstairs{*} Corresponding author: marc.queudot@crim.ca
}
\vspace*{0.1in}
\end{center}

\begin{abstract}
Retrieval-Augmented Generation (RAG) significantly improves the factuality of Large Language Models (LLMs), yet standard pipelines often lack mechanisms to verify intermediate reasoning, leaving them vulnerable to hallucinations in high-stakes domains. To address this, we propose a domain-specific RAG framework that integrates explicit reasoning and faithfulness verification. Our architecture augments standard retrieval with neural query rewriting, BGE-based cross-encoder reranking, and a rationale generation module that grounds sub-claims in specific evidence spans. We further introduce an eight-category verification taxonomy that enables fine-grained assessment of rationale faithfulness, distinguishing between explicit and implicit support patterns to facilitate structured error diagnosis. We evaluate this framework on the BioASQ and PubMedQA benchmarks, specifically analyzing the impact of dynamic in-context learning and reranking under constrained token budgets. Experiments demonstrate that explicit rationale generation improves accuracy over vanilla RAG baselines, while dynamic demonstration selection combined with robust reranking yields further gains in few-shot settings. Using Llama-3-8B-Instruct, our approach achieves 89.1\% on BioASQ-Y/N and 73.0\% on PubMedQA, competitive with systems using significantly larger models. Additionally, we perform a pilot study combining human expert assessment with LLM-based verification to explore how explicit rationale generation improves system transparency and enables more detailed diagnosis of retrieval failures in biomedical question answering.
\end{abstract}

% add your keywords
\copyrightnotice

%{\noindent{\bf Editors:} Lydia Bouzar-Benlabiod, Carson K. Leung}

\section{Introduction}
Large language models (LLMs) have become powerful general-purpose systems for many NLP tasks, including summarization, question answering, translation, code generation, and dialogue. Their progress is largely driven by transformer architectures and large-scale pre-training on diverse corpora \cite{vaswani,brown2020languagemodelsfewshotlearners}. However, LLMs often struggle with factual accuracy, especially in specialized or fast-evolving domains such as medicine, law, or finance \cite{ji2023survey,shuster2021}. Because their knowledge is fixed at pre-training time and limited by the coverage and quality of the training data, they can produce outdated or incorrect outputs when faced with novel or domain-specific information.

Retrieval-augmented generation (RAG) addresses this limitation by combining parametric knowledge in model weights with non-parametric knowledge from external corpora \cite{lewis2020retrieval}. In a RAG pipeline, relevant passages are retrieved from a knowledge base and provided as context to the generator, aiming to ground responses in verifiable evidence and improve factuality and interpretability.

Despite substantial progress, RAG systems face persistent challenges. First, end-to-end performance is highly sensitive to retrieval quality: even small retrieval errors can propagate into generation mistakes \cite{izacard-grave-2021-leveraging}. Second, many RAG systems lack explicit reasoning and verification steps, making them vulnerable to subtle hallucinations (e.g., incorrect dates or conflated entities) even when relevant evidence is retrieved \cite{zhou2024trustworthinessrag}. Third, domain deployments often require specialized taxonomies, lexicons, and continually updated corpora, which generic RAG frameworks under-support and which can lead to degraded reliability out of distribution \cite{singhal2022largelanguagemodelsencode}.

To address these gaps, we propose a domain-specific RAG framework with explicit reasoning and verification, extending \cite{instructrag}. Our approach combines BM25 retrieval with neural query rewriting, BGE-based reranking \cite{bge}, and rationale generation with verification.
\\

In this paper, our main contributions are:
\begin{enumerate}
    \item A reproducible domain-specific RAG blueprint with explicit verification gates. We present a biomedical RAG pipeline that integrates retrieval, reranking, rationale generation, and verification into a modular workflow, and empirically evaluate the impact of reranking and dynamic demonstration selection.
    
    \item A practical, statement-level faithfulness framework for biomedical rationales. We propose and operationalize a verification taxonomy for rationale statements grounded in retrieved abstracts, enabling structured auditing of faithfulness and clearer attribution of errors to retrieval vs. generation.
    
    \item A systematic evaluation of design choices under token and latency constraints. We run controlled experiments isolating the effects of reranking and dynamic demonstration selection. In a preliminary analysis, we use our hybrid verification framework to categorize critical failure modes, distinguishing between implicit and explicit support patterns to inform future rationale generation research.
\end{enumerate}

\section{Related Work}
\subsection{Retrieval-Augmented Generation}
Retrieval-augmented generation (RAG) has emerged as a powerful mechanism for improving the factual accuracy of language models by conditioning them on external corpora at inference time. Early work by ~\cite{lewis2020retrieval} demonstrated that a simple retrieve-and-generate pipeline can significantly boost performance for question-answering tasks by retrieving relevant passages from an external knowledge base. This approach was further refined by introducing joint training of the retriever and generator, yielding improved end-to-end retrieval precision and generation quality ~\cite{izacard-grave-2021-leveraging}. 

\subsection{Explicit Reasoning and Rationale Generation}
Beyond simply retrieving documents, recent efforts have focused on exposing the model's reasoning process to improve transparency and trust. ~\cite{zhou2024trustworthinessrag} proposed TrustworthyRAG, which augments RAG with explicit rationale generation heads that tie each token in the output back to source passages, enabling more precise faithfulness checks. Such structured rationales aid with error analysis and allow downstream modules to verify whether each claim is supported by retrieved evidence.

\subsection{Domain‑Specific Retrieval}
Generic RAG systems often struggle in specialized domains, such as medicine or law, where terminology and factual requirements are more stringent. ~\cite{singhal2022largelanguagemodelsencode} studied the limitations of out‑of‑the‑box LLMs in healthcare applications, showing that domain‑tuned retrievers coupled with domain‑specific corpora can yield substantial gains in both relevance and factuality of generated outputs. Complementary work has investigated query rewriting and reranking strategies to further adapt retrieval to domain‑specific vocabularies and document structures~\cite{ji2023survey}. 

% \subsection{Reasoning-based RAG Systems with Explainability}
% Recent work augments RAG with explicit reasoning to improve transparency and factuality by making intermediate inferences observable and checkable. Two representative directions are (i) rationale-first pipelines that generate and validate evidence-linked explanations before committing to an answer, and (ii) interleaved retrieve–reason loops that expand the context while the model reasons.
% In the first category, \cite{i2023minimizingfactualinconsistencyhallucination} introduce the Factual Evidence (FE) framework, which elicits concise rationales, verifies each rationale against retrieved passages, and refines inconsistencies prior to final answer generation. By decoupling explanation and decision, FE improves faithfulness and traceability, especially in high-stakes domains such as pharmacovigilance, because each sub-claim must be grounded in cited spans of retrieved text before it can influence the answer.
% In the second category, \cite{wang2024ratretrievalaugmentedthoughts} propose Retrieval-Augmented Thoughts (RAT), which integrates step-wise retrieval into chain-of-thought prompting. RAT alternates between generating intermediate thoughts and issuing new queries, enabling multi-hop reasoning when initial evidence is insufficient. This interleaving process reduces hallucinations that arise from reasoning over incomplete context and improves answer quality on compositional queries.

\section{Methodology}
\subsection{Framework Architecture}
We extend the InstructRAG pipeline \cite{instructrag} to target two persistent RAG failure modes: irrelevant retrieval and hallucinated reasoning. We do so by (i) improving the retrieved evidence set presented to the generator and (ii) making the reasoning process explicit, verifiable, and corrigible. An overview of the full control flow is provided in Algorithm~\ref{alg:rvr_rag}, which summarizes retrieval, reranking, optional query rewriting, rationale generation, and statement-level verification. We describe each module and the control logic that ties them together below. 
\\

\paragraph{{\textbf{BM25 Retriever.}}}
We use the BM25 algorithm to quickly gather a broad initial set of potentially relevant documents. Given a user query $q$, the retriever returns the top $k{=}20$ passages ${p_i}_{i=1}^{20}$. BM25's efficiency and robustness to domain-specific lexical cues make it a reliable starting point in specialized corpora.
\\

\paragraph{\textbf{BGE Cross-Encoder Reranker.}}
To enhance evidential precision, we rerank the top 20 passages retrieved by BM25 with a BGE cross-encoder. For each candidate passage $p_i$, the reranker jointly encodes the query--passage pair $(q, p_i)$ and assigns a relevance score that reflects deeper semantic alignment beyond lexical overlap. We then select the top $m{=}5$ passages to form the final evidence set $E$ used in downstream reasoning.
\\

\paragraph{\textbf{Query Rewriter.}}
We use a GPT-4o module to clarify ambiguous queries $q$ by expanding acronyms and adding precise medical terminology. To avoid unnecessary delays, this step is optional: it is only triggered when the initial retrieval results lack significant keyword overlap with the query or when the reranker finds insufficient evidence. 
\\

\paragraph{\textbf{Rationale Generator.}}
We prompt the generator to produce a concise, evidence-linked rationale $R$ alongside a provisional answer $\hat{y}$. Prompts include the retrieved passages $p_i$ and instructions to (i) decompose the question into sub-claims, (ii) cite the specific passage IDs (and if available, character/token spans) that support each sub-claim, and (iii) refrain from using knowledge not present in $p_i$. 
\\

\paragraph{\textbf{Rationale Verifier.}}
Building on the Factual Evidence framework \cite{i2023minimizingfactualinconsistencyhallucination}, we use GPT-4o to classify rationale statements based on their factual alignment with retrieved documents. Each statement is categorized as correct (explicit, implicit, additional info, or missing context) or incorrect (false info, deviating info, illogical, or missing evidence), enabling fine-grained faithfulness assessment.
\\

% \paragraph{\textbf{ReAct Agent.}}
% We implement a ReAct-style agent \cite{yao2023reactsynergizingreasoningacting} that analyzes verification feedback, reformulates queries, retrieves updated passages, and regenerates rationales. The ReAct loop is triggered deterministically whenever the Rationale Verifier assigns an \texttt{INCORRECT} label to the generated rationale.

% \begin{figure}[ht]
%     \centering
%     \includegraphics[width=1.00\linewidth]{custom_architecture.png}
%     \caption{Architecture of the proposed domain-specific RAG framework, featuring explicit rationale generation with an integrated verify-and-refine feedback loop.}
%     \label{fig:extended-architecture}
% \end{figure}

\begin{algorithm}[t]
\caption{Domain-Specific RAG with Explicit Reasoning and Verification}
\label{alg:rvr_rag}
\small
\begin{algorithmic}[1]
\Require user query $q$; corpus index $\mathcal{D}$; BM25 top-$k$; rerank top-$m$;
rewrite threshold $\tau_{\text{ovlp}}$; evidence threshold $\tau_{\text{evid}}$
\Ensure final answer $y$, rationale $R$, evidence set $E$, verification $V$

\State \textbf{Retrieve:} $C \leftarrow \text{BM25}(\mathcal{D}, q, k)$ \Comment{Initial candidate set}
\State \textbf{Rerank:} $E \leftarrow \text{BGE\_Rerank}(q, C, m)$ \Comment{Top-$m$ evidence passages}

\State \textbf{Rewrite trigger:} compute lexical overlap $s \leftarrow \text{Overlap}(q, E)$
\State compute evidence score $e \leftarrow \text{EvidenceScore}(q, E)$
\If{$s < \tau_{\text{ovlp}} \;\lor\; e < \tau_{\text{evid}}$}
    \State $q' \leftarrow \text{Rewrite}(q)$ \Comment{Expand acronyms, add medical terms}
    \State $C \leftarrow \text{BM25}(\mathcal{D}, q', k)$
    \State $E \leftarrow \text{BGE\_Rerank}(q', C, m)$
\EndIf

\State \textbf{Reason:} $(\hat{y}, R) \leftarrow \text{GenerateAnswerAndRationale}(q, E)$
\State \textbf{Verify:} $V \leftarrow \text{VerifyRationale}(q, E, R)$ \Comment{8-category faithfulness labels}

\State \Return $y = \hat{y}$, $R$, $E$, $V$
\end{algorithmic}

\vspace{0.75em}
\hrule
\vspace{0.5em}
\noindent \textbf{\footnotesize Implementation details:} 
\footnotesize 
We set $k{=}20$ and $m{=}5$ in all experiments. 
\textbf{Overlap} computes the fraction of non-stopword query tokens appearing in the concatenated evidence passages. 
\textbf{EvidenceScore} is the mean reranker score of the top-$m$ passages. 
\textbf{GenerateAnswerAndRationale} prompts Llama-3-8B-Instruct with retrieved evidence and (optionally) dynamically selected demonstrations to produce an evidence-linked rationale $R$ and provisional answer $\hat{y}$. 
\textbf{VerifyRationale} uses GPT-4o to label each rationale statement according to the categories in Table~\ref{tab:label_legend_en}.
\end{algorithm}

\subsection{Corpus}
Our knowledge base was sourced from the MedRAG toolkit \cite{medrag}, which includes PubMed abstracts (23M+ biomedical literature abstracts), Wikipedia articles (general medical background), medical textbooks (authoritative exam-style content), and StatPearls (clinical decision support summaries). 

For our experiments, we exclusively used the PubMed abstracts corpus, as both evaluation datasets (BioASQ and PubMedQA) are specifically designed for literature-based question answering and their questions are answerable from scientific abstracts. This focused approach enables direct assessment of our framework's retrieval and reasoning capabilities on domain-appropriate literature without introducing noise from heterogeneous sources.

\subsection{Datasets}
We evaluated our framework on two biomedical QA datasets from the MIRAGE benchmark~\cite{medrag}: BioASQ~\cite{bioasq} and PubMedQA~\cite{jin-etal-2019-pubmedqa}. Both are knowledge-intensive and require domain-specific literature access.

BioASQ contains expert-curated yes/no questions requiring literature-grounded reasoning to verify biomedical claims. PubMedQA provides research questions with three candidate answers (yes/no/maybe) answerable from PubMed abstracts. Following MIRAGE settings, we use question-only retrieval without gold passage supervision, simulating realistic medical information-seeking scenarios.

\subsection{Metrics}
We evaluate our framework using a combination of automatic and human-centric metrics to assess factual accuracy, rationale quality, and inter-annotator agreement for our LLM-as-a-judge components.

For both BioASQ and PubMedQA, we evaluate model performance using classification accuracy. BioASQ is a binary task (\texttt{yes} / \texttt{no}), whereas PubMedQA is a three-way classification (\texttt{yes} / \texttt{no} / \texttt{maybe}).

To measure the alignment between generated rationales and their corresponding retrieved context passages, we use a custom verification schema inspired by the MIRAGE benchmark~\cite{medrag}. Each context document rationale is classified into one of eight categories described in Table~\ref{tab:label_legend_en}. We then assigned a binary \texttt{CORRECT} or \texttt{INCORRECT} verdict for the entire rationale.

For a generated rationale $R$, we segment it into $n$ atomic statements $\{r_j\}_{j=1}^{n}$ and assign each statement a label from Table~\ref{tab:label_legend_en}. We define an \emph{atomic statement} as a single verifiable proposition that can be supported or refuted using the retrieved evidence (e.g., one clinical claim, one association, one numerical outcome). In practice, we segment $R$ by sentence boundaries and further split sentences on clause markers (e.g., ``because'', ``therefore'', ``however'', ``which suggests'') and list delimiters (e.g., semicolons, enumerations) to isolate minimal factual units. We merge fragments that are not semantically self-contained (e.g., pronoun-only continuations) with the preceding clause. This segmentation procedure is deterministic and is applied identically for human annotation and the LLM verifier. We define a binary support indicator $\mathbb{I}_j$ that equals 1 if $r_j$ is labeled as any \texttt{CORRECT-*} category and 0 otherwise (i.e., any \texttt{INCORRECT-*}). The faithfulness score is the proportion of supported statements:
\begin{equation}
\mathrm{Faith}(R) \;=\; \frac{1}{n}\sum_{j=1}^{n} \mathbb{I}_j.
\label{eq:faithfulness}
\end{equation}

To quantify the consistency of rationale annotations, we compute two complementary metrics for each context condition (CONTEXT-1 through CONTEXT-5). CONTEXT-$j$ denotes verification using only the top-$j$ retrieved passages (for $j\in\{1,\dots,5\}$):  

\begin{itemize}
    \item \textbf{Cohen's $\kappa$.} We use Cohen's $\kappa$ to measure chance-corrected agreement. We report $\kappa$ for (i) human--human agreement and (ii) each human annotator versus the LLM verifier. We interpret $\kappa$ using the Landis--Koch guidelines \cite{Landis1977}: slight (0.00--0.20), fair (0.21--0.40), moderate (0.41--0.60), substantial (0.61--0.80), and almost perfect (0.81--1.00).

    \item \textbf{Per-category F$_1$.} For each rationale label category, we compute precision and recall and report the corresponding per-category F$_1$ score. When comparing humans to the LLM verifier, we treat the human label as the reference and compute F$_1$ for the verifier on each category.
\end{itemize}

By reporting both $\kappa$ and F$_1$ for human–human and human–model pairs under every context, we can obtain a nuanced view of annotation reliability and model alignment.

\begin{table*}[t]
\centering
\caption{Legend for Rationale-Verification Label Categories}
\footnotesize
\setlength{\tabcolsep}{6pt}
\renewcommand{\arraystretch}{1.1}
\newcolumntype{Y}{>{\raggedright\arraybackslash}X}
\begin{tabularx}{\textwidth}{lY}
\toprule
\textbf{Category} & \textbf{Meaning} \\ \midrule
\texttt{CORRECT-EXPLICIT} &
Information is explicitly stated in the documents (quoted or paraphrased). \\
\texttt{CORRECT-IMPLICIT} &
Facts are not stated verbatim but are logically inferred from context clues. \\
\texttt{CORRECT-ADDITIONAL} &
Uses context accurately but adds relevant, correct external details. \\
\texttt{CORRECT-MISSING} &
Conclusion is correct, but the cited documents provide no support (irrelevant). \\
\texttt{INCORRECT-FALSE} &
Statements directly contradict evidence provided in the context. \\
\texttt{INCORRECT-DEVIATING} &
Statements are off-topic or unrelated to the query/documents. \\
\texttt{INCORRECT-ILLOGICAL} &
Reasoning contains internal contradictions or violates logic/scientific principles. \\
\texttt{INCORRECT-MISSING} &
Reasoning is incorrect \emph{and} the cited documents are irrelevant. \\
\bottomrule
\end{tabularx} 
\label{tab:label_legend_en}
\end{table*}

\section{Experiments}

\subsection{Experimental Design}

Our experimental framework is designed to isolate and evaluate the specific contributions of retrieval, reranking, and explicit reasoning verification in biomedical question answering. We conduct a series of controlled experiments to assess system performance under varying constraints.
\\

\paragraph{\textbf{Demonstration Pool Construction ($T^{\star}$).}}
We construct a static pool of in-context demonstrations, $T^{\star}$, \emph{offline} using exclusively the training split of each dataset. For each training example $(q_i, y_i)$, we retrieve evidence passages from the same PubMed index used at test time and generate an evidence-linked rationale $r_i$ using the same prompt format as in evaluation. Each demonstration is stored as a tuple $(q_i, r_i, y_i, E_i)$, where $E_i$ denotes the set of top-$m$ retrieved passages. Crucially, the rationales in $T^{\star}$
 are model-generated rather than human-curated; this ensures that demonstrations reflect the model's achievable reasoning patterns rather than potentially unattainable gold-standard explanations.
\\

\paragraph{\textbf{Dynamic Demonstration Selection.}}
At inference time, we select the top-$k$ demonstrations from $T^{\star}$ based on cosine similarity between the embedding of the evaluation query and the stored training-query embeddings. Our primary policy is similarity-based; however, we note the risk of label-prior bias, particularly for the ternary classification in PubMedQA (\emph{yes/no/maybe}). We therefore evaluate both (i) a strict \textit{similarity-only} policy and (ii) a \textit{label-balanced} policy that enforces class diversity among the retrieved demonstrations. Unless otherwise stated, we report results under the similarity-only setting.
\\

\paragraph{\textbf{Data Decontamination.}}
To ensure robust evaluation, we enforce strict decontamination protocols. The demonstration pool $T^{\star}$ is disjoint from all validation and test splits. We further mitigate leakage by performing embedding-based deduplication to remove near-duplicate questions across splits. Additionally, when document identifiers are available (e.g., PubMed IDs), we filter out any demonstration that shares an identifier with the current evaluation query, preventing the model from exploiting memorized document--answer associations. Random seeds are fixed for all fallback sampling procedures.

\subsubsection{Comparative Approaches} \label{comp approaches} We evaluate the proposed framework against the following configurations:

\begin{itemize} 
    \item \textbf{Vanilla RAG}: A baseline that retrieves documents using BM25 and generates answers directly, lacking an explicit intermediate reasoning step.
    
    \item \textbf{InstructRAG-ICL}: Our proposed baseline that incorporates a distinct reasoning generation step via in-context learning prior to producing the final answer. This entails using examples of similar tasks (demonstrations).
    
    \item \textbf{InstructRAG w/ Reranker}: An enhancement of the ICL approach that applies a BGE-v2-m3 cross-encoder to rerank the initial retrieval candidates, improving the precision of the evidence $E$.
    
    \item \textbf{InstructRAG w/ Reranker + Dynamic ICL}: Extends the reranker model by replacing static demonstrations with dynamically selected examples via $k$-nearest neighbor (KNN) search.
\end{itemize}

% \begin{itemize} 
%     \item \textbf{Vanilla RAG}: A baseline that retrieves documents using BM25 and generates answers directly, lacking an explicit intermediate reasoning step.
    
%     \item \textbf{InstructRAG-ICL}: Our proposed baseline that incorporates a distinct reasoning generation step via in-context learning prior to producing the final answer.
    
%     \item \textbf{InstructRAG w/ Reranker}: An enhancement of the ICL approach that applies a BGE cross-encoder to rerank the initial retrieval candidates, improving the precision of the evidence $E$.
    
%     \item \textbf{InstructRAG w/ Reranker + Dynamic ICL}: Extends the reranker model by replacing static demonstrations with dynamically selected examples via $k$-nearest neighbor (KNN) search.
    
%     \item \textbf{InstructRAG w/ Reranker + Dynamic ICL + ReAct}: The complete proposed framework (``Reason-Verify-Refine''), incorporating a ReAct agent that triggers a rewrite-and-retrieve loop when the verification module deems the initial rationale insufficient.
    
%     \item \textbf{InstructRAG w/ Reranker + Dynamic ICL + o3}: A variant replacing the Llama-3-8B-Instruct backbone with OpenAI's o3 model to assess the impact of a stronger underlying reasoning engine.
% \end{itemize}

\subsubsection{Number Of Demonstrations}
To understand the impact of demonstration examples on reasoning quality and overall performance, we tested configurations ranging from 0 to 4 samples for few-shot learning to evaluate the trade-off between demonstration quantity and context window utilization.

For each evaluation query, we (i) computed cosine similarity between the query vector and all training vectors, (ii) selected the top-k most similar examples while enforcing a diversity constraint, and (iii) ensured a minimum unique example count through random selection if needed.

\subsubsection{Query Rewriting}
The optional query rewriting module is triggered when lexical overlap falls below $\tau_{\text{ovlp}} = 0.3$ or when the mean reranker score is below $\tau_{\text{evid}} = 0.5$. These thresholds were set once based on manual inspection of 50 training queries and were not tuned further. In our evaluation, query rewriting was triggered in approximately 8\% of BioASQ queries and 12\% of PubMedQA queries. We did not ablate the rewriting component in isolation, as our primary focus was on reranking and demonstration selection; a systematic study of query rewriting strategies and threshold sensitivity is left for future work.

\subsubsection{Reranking Experiments}
We evaluated the impact of the BGE reranker by comparing performance across all ICL variations both with and without the reranking component. For the reranking experiments, we (i) retrieved an initial set of 20 candidate documents using BM25, (ii) applied the BGE cross-encoder to rerank these candidates, and (iii) selected the top-5 documents after reranking for rationale generation.

We chose 20 initial candidates and kept the top-5 after reranking as an empirically validated, budget-aware setting that matches common two-stage retrieval practice and preserved our token/latency constraints.

\subsubsection{Retrieval Method}
For document retrieval, we initially used the MedCPT retriever. However, to simplify our experimental pipeline and conserve computational resources, we transitioned to BM25. This change was supported by similar performance metrics between BM25 and more complex retrievers in our preliminary evaluations. 

\subsubsection{Model}
For all experiments, we used Llama-3-8B-Instruct for both reasoning generation (in the case of InstructRAG-ICL) and answer generation (for both InstructRAG-ICL and Vanilla RAG) (see subsection \ref{comp approaches})

\subsection{Human Evaluation Setup}

To assess the quality of generated rationales and validate our automatic evaluation metrics, we designed a human annotation study with the following components:

\subsubsection{Annotation Protocol}
Two annotators with extensive experience in NLP systems independently evaluated a convenience sample of 4 examples (2 per dataset). While this sample size precludes statistical inference, it allowed us to (i) validate the clarity of our 8-category taxonomy, (ii) identify systematic differences between human and LLM judgments, and (iii) surface failure modes for qualitative analysis. For each example, annotators evaluated:

\begin{itemize}
    \item \textbf{Overall Response Quality}: A 1-5 scale rating the comprehensiveness and correctness of the final answer.
    \item \textbf{Rationale-Context Alignment}: Classifying each reasoning statement according to its faithfulness to the retrieved documents using eight predefined categories.
    \item \textbf{Faithfulness Score}: A numeric score on how much the response is grounded in the provided context.
\end{itemize}

\subsubsection{LLM-as-a-Judge Implementation}
In parallel with human annotation, we implemented an automated evaluation using a large language model (GPT-4o) to assess the system output on the same examples used for the human evaluation. The automated judge was provided with: (i) the original question, (ii) retrieved context documents, (iii) the generated rationale, and (iv) the generated answer. The LLM judge was instructed to evaluate using the same criteria as human annotators.

\section{Results}

We present our experimental findings in three parts: (1) comparison with published baselines, (2) ablation analysis of our framework components, and (3) analysis of demonstration selection strategies.

\subsection{Comparison with Baselines}

Table~\ref{tab:main_comparison} compares our configurations against 
published MIRAGE baselines. Despite using Llama-3-8B-Instruct—a model 
roughly 10$\times$ smaller than GPT-4—our framework achieves competitive 
or superior performance.

On BioASQ-Y/N, our best configuration (3-shot Dynamic ICL with reranking) 
achieves 89.1\% accuracy, approaching MedRAG+GPT-3.5 (90.29\%). On 
PubMedQA*, our 0-shot rationale generation achieves \textbf{73.0\%}, 
exceeding MedRAG+GPT-4 (70.60\%) by 2.4 points absolute. We attribute 
this to explicit rationale generation: by requiring the model to 
articulate evidence-linked reasoning before answering, we reduce 
hallucinated inferences that may arise in larger models operating 
without such structure.

Notably, while MedRAG uses a four-corpus ensemble (MedCorp) with 
reciprocal rank fusion over four retrievers (RRF-4), our system 
uses only PubMed with BM25 + BGE reranking—a substantially simpler 
retrieval pipeline. This suggests that reasoning-side improvements 
(rationale generation, dynamic ICL) can compensate for retrieval 
complexity.

\begin{table}[t]
\centering
\caption{Comparison with baseline methods under MIRAGE settings. Accuracy (\%) reported on \textbf{BioASQ-Y/N} (618 yes/no questions) and \textbf{PubMedQA*} (500 questions, context removed). All baselines use question-only retrieval without gold passages. $^\dagger$Results from MIRAGE benchmark~\cite{medrag}.}
\label{tab:main_comparison}
\small
\setlength{\tabcolsep}{5pt}
\renewcommand{\arraystretch}{1.12}
\begin{tabular}{lcc}
\toprule
\textbf{Method} & \textbf{BioASQ-Y/N} & \textbf{PubMedQA*} \\
\midrule
\multicolumn{3}{l}{\textit{Closed-Book (No Retrieval; CoT Prompting)}} \\
\quad GPT-3.5$^\dagger$ & 74.27 & 36.00 \\
\quad GPT-4$^\dagger$ & 84.30 & 39.60 \\
\quad Mixtral-8$\times$7B$^\dagger$ & 77.51 & 35.20 \\
\quad Llama2-70B$^\dagger$ & 61.17 & 42.20 \\
\quad MEDITRON-70B$^\dagger$ & 68.45 & 53.40 \\
\quad PMC-LLaMA-13B$^\dagger$ & 63.11 & 55.80 \\
\midrule
\multicolumn{3}{l}{\textit{MedRAG (MedCorp + RRF-4 Retriever)}} \\
\quad + GPT-3.5$^\dagger$ & \underline{90.29} & 67.40 \\
\quad + GPT-4$^\dagger$ & \textbf{92.56} & 70.60 \\
\quad + Mixtral-8$\times$7B$^\dagger$ & 87.54 & 67.60 \\
\quad + Llama2-70B$^\dagger$ & 73.95 & 50.40 \\
\quad + MEDITRON-70B$^\dagger$ & 76.86 & 56.40 \\
\midrule
\multicolumn{3}{l}{\textit{MedRAG Variants (GPT-3.5 backbone)}} \\
\quad PubMed + BM25$^\dagger$ & 88.51 & 66.20 \\
\quad PubMed + MedCPT$^\dagger$ & 85.76 & 66.40 \\
\quad MedCorp + BM25$^\dagger$ & 87.70 & 66.20 \\
\quad MedCorp + RRF-2$^\dagger$ & 88.19 & 67.80 \\
\midrule
\multicolumn{3}{l}{\textit{Ours (Llama-3-8B-Instruct backbone)}} \\
\quad Vanilla RAG (BM25) & 82.3 & 70.0 \\
\quad + Rationale Gen.\ (0-shot) & 85.8 & \textbf{73.0} \\
\quad + Reranking (0-shot) & 87.4 & \underline{72.5} \\
\quad + Dynamic ICL (best-$k$) & 89.1 & 71.0 \\
\bottomrule
\end{tabular}
\vspace{0.3em}

\raggedright
\footnotesize
\textbf{Notes:} Best-$k$ = 3-shot for BioASQ-Y/N, 2-shot for PubMedQA*. Best results per section in \textbf{bold}; second best \underline{underlined}. MedCorp combines PubMed, StatPearls, Textbooks, and Wikipedia corpora. RRF-$n$ denotes Reciprocal Rank Fusion over $n$ retrievers.
\end{table}

\subsection{Ablation Study}

Table~\ref{tab:ablation_study} isolates the contribution of each framework component. We report results across different numbers of in-context demonstrations to understand interactions between components.

\begin{table*}[t]
\centering
\caption{Ablation study showing the impact of reranking and demonstration selection strategy. Accuracy (\%) reported across varying numbers of ICL demonstrations.}
\label{tab:ablation_study}
\small
\setlength{\tabcolsep}{4pt}
\renewcommand{\arraystretch}{1.1}
\begin{tabular}{lccccc|ccccc}
\toprule
& \multicolumn{5}{c|}{\textbf{BioASQ}} & \multicolumn{5}{c}{\textbf{PubMedQA}} \\
\cmidrule(lr){2-6} \cmidrule(lr){7-11}
\textbf{Configuration} & \textbf{0} & \textbf{1} & \textbf{2} & \textbf{3} & \textbf{4} & \textbf{0} & \textbf{1} & \textbf{2} & \textbf{3} & \textbf{4} \\
\midrule
\multicolumn{11}{l}{\textit{Static Demonstration Selection}} \\
\quad w/o Reranking & 85.8 & 78.9 & 78.5 & 77.7 & 70.4 & 73.0 & 54.0 & 60.0 & 56.0 & 47.5 \\
\quad w/ Reranking & 87.4 & 79.8 & 79.6 & 76.4 & 71.7 & 72.5 & 60.5 & 60.0 & 60.0 & 60.0 \\
\quad $\Delta$ (Reranking) & \textcolor{green!50!black}{+1.6} & \textcolor{green!50!black}{+0.9} & \textcolor{green!50!black}{+1.1} & \textcolor{red}{-1.3} & \textcolor{green!50!black}{+1.3} & \textcolor{red}{-0.5} & \textcolor{green!50!black}{+6.5} & -- & \textcolor{green!50!black}{+4.0} & \textcolor{green!50!black}{+12.5} \\
\midrule
\multicolumn{11}{l}{\textit{Dynamic Demonstration Selection (w/ Reranking)}} \\
\quad Dynamic ICL & 87.4 & 88.7 & 88.3 & \textbf{89.1} & 86.2 & 72.5 & 65.0 & \textbf{71.0} & 66.5 & 69.0 \\
\quad $\Delta$ vs.\ Static & -- & \textcolor{green!50!black}{+8.9} & \textcolor{green!50!black}{+8.7} & \textcolor{green!50!black}{+12.7} & \textcolor{green!50!black}{+14.5} & -- & \textcolor{green!50!black}{+4.5} & \textcolor{green!50!black}{+11.0} & \textcolor{green!50!black}{+6.5} & \textcolor{green!50!black}{+9.0} \\
\bottomrule
\end{tabular}
\vspace{0.3em}

\raggedright
\footnotesize
\textbf{Note:} 0-shot results are identical for static and dynamic selection (no demonstrations used). $\Delta$ shows improvement from the ablated component. Best results per dataset in \textbf{bold}.
\end{table*}

\subsubsection{Effect of Reranking}

Reranking with the BGE cross-encoder provides consistent benefits, particularly for PubMedQA in few-shot settings. The most dramatic improvement occurs in the 4-shot PubMedQA setting, where reranking improves accuracy from 47.5\% to 60.0\% (+12.5 points). This suggests that reranking helps filter out noisy passages that would otherwise mislead the model when combined with multiple demonstrations.

For BioASQ, reranking yields modest but consistent gains in the 0--2 shot range (+0.9 to +1.6 points). The slight degradation at 3-shot (-1.3 points) may indicate that high-quality demonstrations can partially compensate for retrieval noise.

\subsubsection{Effect of Dynamic Demonstration Selection}

Dynamic ICL selection via KNN retrieval substantially outperforms static demonstration selection across all few-shot configurations. The improvements are particularly pronounced for BioASQ, where dynamic selection at 4-shot yields +14.5 points over static selection (86.2\% vs.\ 71.7\%).

Notably, dynamic selection reverses the degradation pattern observed with static ICL. Under static selection, adding more demonstrations \emph{hurts} performance on both datasets (BioASQ drops from 85.8\% at 0-shot to 70.4\% at 4-shot). With dynamic selection, performance remains stable or improves, peaking at 3-shot for BioASQ (89.1\%) and 2-shot for PubMedQA (71.0\%).

\subsection{Analysis of Demonstration Sensitivity}

PubMedQA exhibits higher sensitivity to demonstration selection than BioASQ. We attribute this to two factors: (i) the ternary classification structure (yes/no/maybe) creates label-prior bias when demonstrations are not carefully balanced, and (ii) PubMed abstracts are longer, causing additional demonstrations to compete with retrieved evidence for context window space.

\section{Limitations}

Despite promising results on biomedical question answering, our study has several limitations. First, evaluation is limited to two English biomedical datasets (BioASQ and PubMedQA), which may reduce generalizability to other domains, languages, or question formats. Both benchmarks largely emphasize factoid and binary judgments; performance on more complex settings (e.g., multi-hop or causal reasoning) remains untested.

Second, parts of our pipeline rely on OpenAI API calls, introducing additional latency and monetary cost that may hinder deployment at scale. Exploring open-source substitutes for these components is an important direction for future work.

Third, we do not evaluate in real clinical workflows or with clinicians in the loop. Our system is intended as a research prototype for studying rationale generation and verification rather than a clinical decision support tool. Any high-stakes deployment would require substantially more validation and safety measures.

Fourth, we report single-point accuracy estimates without statistical significance testing or confidence intervals. While our ablation design isolates component contributions, the magnitude of improvements should be interpreted with appropriate caution until replicated across multiple runs.

Finally, our human evaluation is small (4 examples annotated by 2 raters). Larger studies are needed to reliably assess rationale quality, agreement, and the practical utility of explicit reasoning for biomedical QA.

\section{Demo}
A functional demonstration of the proposed system is available on HuggingFace\footnote{\url{https://huggingface.co/spaces/DialogueRobust/RobustDialogueDemo}}.
In this interactive application, users can query an AI-related white paper, and the system responds with both a generated answer and a transparent rationale. Specifically, the interface displays the supporting evidence passages retrieved from the document collection and explains how they were integrated into the model's response. This allows users to assess the relevance of the retrieved information and the robustness of the underlying retrieval-augmented generation pipeline.

% \section{\hl{Acknowledgments -- À COMPLÉTER}}

\appendix
\section{Pilot Human Study: Per-Example Faithfulness Scores}
\label{app:pilot_faithfulness}

As an illustrative complement to our pilot human evaluation (4 examples total), Table~\ref{tab:pilot_faithfulness_scores} reports the per-example faithfulness scores $\mathrm{Faith}(R)$ (Eq.~\ref{eq:faithfulness}) assigned by two human annotators and the LLM verifier. We include these values to qualitatively examine agreement patterns and to illustrate cases where the automated verifier appears more strict or more permissive than the human raters. Given the small sample size, these results are descriptive and should not be interpreted as statistically reliable estimates of verifier performance.

\begin{table}[h]
\centering
\caption{Pilot per-example faithfulness scores (4 examples). Scores are shown for two human annotators and the LLM verifier.}
\label{tab:pilot_faithfulness_scores}
\small
\setlength{\tabcolsep}{8pt}
\renewcommand{\arraystretch}{1.15}
\begin{tabular}{lccc}
\toprule
\textbf{Example} & \textbf{Annotator A} & \textbf{Annotator B} & \textbf{LLM Verifier} \\
\midrule
Question 1 & 1.00 & 1.00 & 1.00 \\
Question 2 & 0.75 & 0.50 & 0.83 \\
Question 3 & 0.75 & 0.30 & 0.93 \\
Question 4 & 0.90 & 0.80 & 1.00 \\
\bottomrule
\end{tabular}
\end{table}

The LLM verifier tends to assign higher faithfulness scores than human annotators (mean 0.94 vs.\ 0.85 and 0.65), suggesting it may be more permissive in accepting implicit reasoning. Human annotators show substantial disagreement on PubMedQA-1 (0.75 vs.\ 0.30), highlighting the subjectivity inherent in faithfulness assessment and the need for clearer annotation guidelines in future work.

\section{Prompts and Instructions}
\label{app:prompts}

For reproducibility, we report the prompt templates used in our framework. Prompts differ slightly across datasets due to output format requirements (BioASQ vs.\ PubMedQA).

\subsection{Base Instructions}
The following system instruction was prepended to all queries.

\begin{quote}
\small
\textbf{Task Instruction:} Analyze the provided documents and answer the question. Briefly explain how the documents support your answer. If the documents are not useful, answer from your own knowledge without referencing them.
\end{quote}

\subsection{Dataset-Specific Instructions}

\subsubsection{PubMedQA Prompt}
PubMedQA requires ternary classification (yes/no/maybe). We appended:

\begin{quote}
\small
Critically evaluate the medical evidence in the documents (methods, sample sizes, statistical significance). Weigh supporting and opposing evidence, note limitations, and provide concise reasoning followed by a final judgment.

\textbf{OUTPUT FORMAT REQUIREMENT:}
End your response with exactly one of the following on a new line:
``FINAL ANSWER: A. yes'' \\
``FINAL ANSWER: B. no'' \\
``FINAL ANSWER: C. maybe''

If no document supports an answer, output: ``ANSWER UNAVAILABLE''
\end{quote}

\subsubsection{BioASQ Prompt}
BioASQ requires binary classification (yes/no). We appended:

\begin{quote}
\small
Provide a precise factual answer grounded in the documents. Extract relevant statements and justify the decision based strictly on presented facts.

\textbf{OUTPUT FORMAT REQUIREMENT:}
End your response with one of the following on a new line:
``FINAL ANSWER: A. yes'' \\
``FINAL ANSWER: B. no''

If no document supports an answer, output: ``ANSWER UNAVAILABLE''
\end{quote}

% All references should be stored in the file "references.bib".
% That call to use that file is in "cai.cls". 
% Please do not modify anything below this line.
\printbibliography[heading=subbibintoc]

\end{document}